\documentclass[runningheads]{llncs}
\usepackage[colorlinks]{hyperref}
\usepackage[T1]{fontenc}
\usepackage{graphicx}
\usepackage{multirow}
\usepackage{amsfonts,amssymb}
\usepackage{booktabs}
\usepackage{bm}
\usepackage{makecell}
\usepackage{siunitx}
\usepackage{marvosym}

\begin{document}
\title{SFusion: Self-attention based N-to-One Multimodal Fusion Block}
\titlerunning{SFusion}
%
\author{Zecheng Liu\inst{1} \and
	Jia Wei\inst{1}\textsuperscript{(\Letter)} \and
	Rui Li\inst{2} \and Jianlong Zhou\inst{3}}

\authorrunning{Z. Liu et al.}
%
\institute{School of Computer Science and Engineering, South China University of Technology, Guangzhou, China\\
	\email{msaiyan@mail.scut.edu.cn, csjwei@scut.edu.cn}\\
	\and
	Golisano College of Computing and Information Sciences, Rochester Institute of Technology, Rochester, NY, USA\\
	\email{rxlics@rit.edu}
	\and
	Data Science Institute, University of Technology Sydney, Ultimo, NSW 2007, Australia\\
	\email{jianlong.zhou@uts.edu.au}}
\maketitle              
\begin{abstract}
People perceive the world with different senses, such as sight, hearing, smell, and touch. Processing and fusing information from multiple modalities enables Artificial Intelligence to understand the world around us more easily. However, when there are missing modalities, the number of available modalities is different in diverse situations, which leads to an N-to-One fusion problem. To solve this problem, we propose a self-attention based fusion block called SFusion. Different from preset formulations or convolution based methods, the proposed block automatically learns to fuse available modalities without synthesizing or zero-padding missing ones. Specifically, the feature representations extracted from upstream processing model are projected as tokens and fed into self-attention module to generate latent multimodal correlations. Then, a modal attention mechanism is introduced to build a shared representation, which can be applied by the downstream decision model. The proposed SFusion can be easily integrated into existing multimodal analysis networks. In this work, we apply SFusion to different backbone networks for human activity recognition and brain tumor segmentation tasks. Extensive experimental results show that the SFusion block achieves better performance than the competing fusion strategies. Our code is available at \url{https://github.com/scut-cszcl/SFusion}.

\end{abstract}
\section{Introduction}
People perceive the world with signals from different modalities, which often carry complementary information about varying aspects of an object or event of interest. Therefore, collecting and utilizing multimodal information is crucial for Artificial Intelligence to understand the world around us. Data collected from various sensors (e.g., microphones, cameras, motion controllers) are used to identify human activity \cite{cjk:15}. Moreover, multimodal medical images obtained from different scanning protocols (e.g., Computed Tomography, Magnetic Resonance Imaging) are employed for disease diagnosis \cite{glh:19}. Satisfactory performances have been achieved with these multimodal data.

\begin{figure*}[t]
	\centerline{\includegraphics[width=\textwidth]{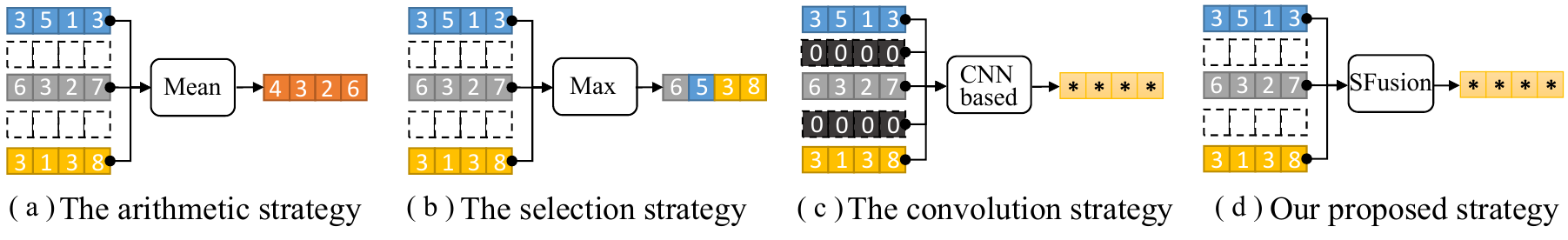}}
	\caption{Fusion strategies. \textbf{$ \ast $} denotes the value is automatically learned.}
	\label{fig1}
\end{figure*}

In practical application, however, modality missing is a common scenario. Wirelessly connected sensors may occasionally disconnect and temporarily be unable to send any data \cite{csc:13}. Medical images may be missing due to artifacts and diverse patient conditions \cite{gm:13}. In these unexpected situations, any combinatorial subset of available modalities can be given as input. To handle this, one intuitive solution is to train a dedicated model on all possible subsets of available modalities \cite{cdjlh:21,hmz:20,wzl:21}. However, these methods are ineffective and time-consuming. Another way is to predict missing modalities and perform with the completed modalities \cite{szw:21}. But, these approaches also require additional prediction networks for each missing situation, and the quality of the recovered data directly affects the performance, especially when there are only a few available modalities. Recently, fusing the available modalities into a shared representation received wide attention. However, it is particularly challenging due to the varying number of input modalities, which results in the N-to-One fusion problem.

Currently, existing fusion strategies to tackle this challenge can be broadly grouped into three categories: the arithmetic strategy, the selection strategy and the convolution strategy. As shown in Fig.~\ref{fig1}(a), in the arithmetic strategy, feature representations of available modalities are merged by an arithmetic function, such as averaging, computing the first and second moments or other designed formulas \cite{djm:19,las:19,hgc:16}. For the selection strategy, as shown in Fig.~\ref{fig1}(b), each value of fused representation is selected from the values at the corresponding position of the inputs. The selection rule can be defined as max, min or probability-based \cite{cjg:18,cl:19,oap:21}. Although the above two fusion strategies are easily scalable to various data missing situations, their fusion operation is hard-coded. All available modalities contribute equally and their latent correlations are neglected. Unlike hard-coding the fusion operation, in the convolution strategy, the convolutional fusion network automatically learns how to fuse these feature representations, which is beneficial to exploiting the correlation between multiple modalities. However, as shown in Fig.~\ref{fig1}(c), this fusion strategy needs a constant number of data to meet the requirements of the input channels in the convolutional network. Therefore, it has to simulate missing data by crudely zero-padding or replacing it with similar modalities, which inevitably introduces a bias in computation and causes performance degradation \cite{cdj:19,nkk:11,lcr:21}.

Transformer has achieved success in the field of computer vision, demonstrating that self-attention mechanism has the ability to capture the latent correlation of image tokens. However, no work has explored the effectiveness of self-attention mechanism on the N-to-One fusion, where N is variable during training, rather than fixed. Furthermore, the calculation of self-attention does not require a fixed number of tokens as input, which represents a potential for handling missing data. Therefore, we propose a self-attention based fusion block (SFusion) to tackle the problems of the above fusion strategies. As shown in Fig.~\ref{fig1}(d), SFusion can handle any number of input data instead of fixing its number. In addition, SFusion is a learning-based fusion strategy that consists of two components: the correlation extraction (CE) module and the modal attention (MA) module. In the CE module, feature representations extracted from available modalities are projected as tokens and fed into the self-attention layers to learn multimodal correlations. Based on these correlations, a modal softmax function is proposed to generate weight maps in the MA module. Finally, it builds a shared feature representation by fusing the varying inputs with the weight maps.

The contributions of this work are:
\begin{itemize}
	\item We propose SFusion, which is a data-dependent fusion strategy without impersonating missing modalities. It can learn the latent correlations between different modalities and builds a shared representation adaptively. 
	
	\item The SFusion is not limited to specific deep learning architectures. It takes inputs from any kind of upstream processing model and serves as the input of the downstream decision model, which enables applying the SFusion to various backbone networks for different tasks.
	
	\item  We provide qualitative and quantitative performance evaluations on activity recognition with the SHL~\cite{shl:19} dataset and brain tumor segmentation with the BraTS2020~\cite{brats:20} dataset. The results show the superiority of SFusion over competing fusion strategies.
	
\end{itemize}

\begin{figure*}[t]
	\centering
	\includegraphics[width=0.9\textwidth]{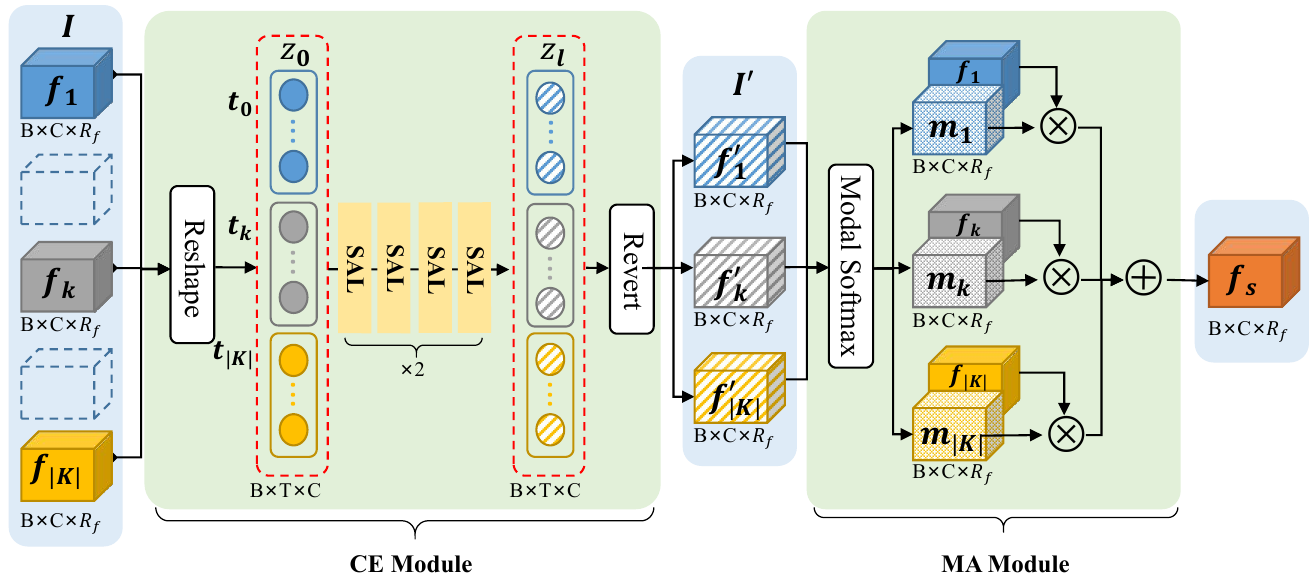}
	\caption{The illustration of SFusion. $ R_{f} $ : L or H$ \times $W or D$ \times $H$ \times $W (shape of feature representation); T = L$ \cdot|K| $ or H$ \cdot $W$ \cdot |K|$ or D$ \cdot $H$ \cdot $W$ \cdot |K|$ (number of tokens).
	}
	\label{fig2}
\end{figure*}

\section{Methodology}

\subsection{Method Overview}

For multiple modalities, let $ k \in K \subseteq \{1,2,\ldots,S\} $ index a specific modality, within the available modality set of $ K $, where $ S $ is the number of all possible modalities. Given an input $ f_{k} \in \mathbb{R}^{B \times C \times R_{f} }$, B and C denote the batch size and the number of channels, respectively. $ R_{f} $ represents the shape of feature representation extracted from the $ k $-th modality of a sample data, which can be 1D (L), 2D (H$ \times $W), 3D (D$ \times $H$ \times $W) or higher-dimensional. In addition, $ I = \{f_{k}|k \in K\} $ denotes the input set of feature representations from all the available modalities. Our goal is to learn a fusion function $ F $ that can project $ I $ into a shared feature representation $ f_{s} $, denoted as $ F(I) \to f_{s} $. To achieve the goal, we design an N-to-One fusion block, SFusion. The architecture is shown in Fig.~\ref{fig2}, which consists of two modules: correlation extraction (CE) module and modal attention (MA) module.
\subsection{Correlation Extraction}
Given the feature representation $ f_{k} \in \mathbb{R}^{B \times C \times R_{f} }$, we first flatten the $ R_{f} $ dimensions of $ f_{k} $ into one dimension and get a $ B \times C \times R$ feature representation, where $ R=L$ (1D), $ R=H \times W $ (2D), $ R=D \times H \times W $ (3D), etc. It can be viewed as $B \times R $ $ C $-dimensional tokens $ t_{k} $. Then, we obtain the concatenation of all the tokens $ z_{0} \in \mathbb{R}^{B \times T \times C} $, where $ T=R\times |K| $, and $ |K| $ denotes the number of available modalities. 

Given $ z_{0} $, the stack of eight self-attention layers (SAL) are introduced to learn the latent multimodal correlations. Each layer includes a multi-head attention (MHA) block and a fully connected feed-forward network (FFN) \cite{vsp:17}. Layer normalization (LN) is applied before every block. The outputs of the $ x $-th ($ x \in [1,2,\ldots,8] $) layer can be describe as:
\begin{equation}
	z^{\prime}_{x} = MHA(LN(z_{x-1}))+z_{x-1}
	\label{eq:transformer1}
\end{equation}
\begin{equation}
	z_{x} = FFN(LN(z^{\prime}_{x}))+z^{\prime}_{x}
	\label{eq:transformer2}
\end{equation}
Therefore, we get $ z_{l} \in \mathbb{R}^{B \times T \times C} $, which is the last SAL output. By reverting $ z_{l} $ to the size of $ |K|\times B \times C \times R_{f} $, we obtain the output $ I^{\prime} = \{f^{\prime}_{k}|k \in K\} $ of CE as:
\begin{equation}
	I^{\prime} = split(r(z_{l}))
	\label{eq:reprojection1}
\end{equation}
where $ r(\cdot) $ and $ split(\cdot) $ are the reshape and split operations, and $ I^{\prime} $ is the set of calculated feature representations $ f^{\prime}_{k} \in \mathbb{R}^{B \times C \times R_{f}} $ which contains multimodal correlations and has the same size as the original input  $ f_{k}$.


\subsection{Modal Attention}
\begin{figure}[b]
	\begin{minipage}[t]{0.45\textwidth}
		\vspace{12pt}
		\centering
		\includegraphics[width=\textwidth]{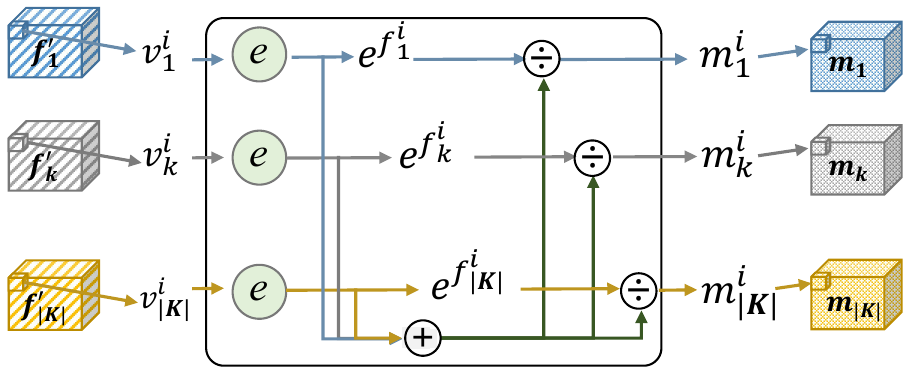}
		\caption{The illustration of modal attention mechanism.}
		\label{fig3}
	\end{minipage}
	\begin{minipage}[t]{0.55\textwidth}
		\vspace{0pt}
		\centering
		\scriptsize 
		\renewcommand{\arraystretch}{1.1}
		\makeatletter\def\@captype{table}\makeatother\caption{Evaluation on SHL2019. $ w/o $ means without. $ \dag $ denotes results from \cite{cl2:19}.}
		\setlength{\tabcolsep}{0.4mm}{\begin{tabular}{l|ccccc}
				\hline
				Accuracy(\%) & Bag & Hips & Torso & Hand & All \\
				\hline
				Early$ \dag $ & -- & -- & -- & -- & 46.73  \\
				Intermediate$ \dag $ & -- & -- & -- & -- & 63.87  \\
				Late$ \dag $ & -- & -- & -- & -- & 63.85  \\
				Confidence$ \dag $\cite{cl:18} & -- & -- & -- & -- & 63.60  \\
				EmbraceNet$ \dag $ \cite{cl2:19} & 63.68 & 67.98 & 81.58 & 47.63 & 65.22  \\
				SFusion & \textbf{67.41} & \textbf{68.91} & \textbf{85.22} & \textbf{48.35} & \textbf{67.47}  \\
				\hline
				SFusion $ w/o $ CE & 56.82&	63.14&	74.69&	46.70&	60.33  \\
				SFusion $ w/o $ MA & 65.01&	67.95&	83.49&	47.52&	65.99  \\
				\hline
		\end{tabular}}
		\label{tab3}%
	\end{minipage}
\end{figure}

Given the calculated feature representations set $ I^{\prime} $, the weight map $ m_{k} $ is generated with the modal attention mechanism. Feature representations extracted from different modalities are expected to have different weights for fusion at the voxel level. Therefore, we introduce a modal-wise and voxel-level softmax function to generate the weight maps from $ I^{\prime} $, as shown in Fig.~\ref{fig3}. 

We denote the $ i $-th voxel of $ f^{\prime}_{k} $ and $ m_{k} $ as $ v^{i}_{k} $ and $ m^{i}_{k} $, respectively. $ e $ is the natural logarithm. The value of weight map $ m_{k} $ can be defined as:
\begin{equation}
	m^{i}_{k} = \frac{e^{v^{i}_{k}}}{\sum_{j \in K}{e^{v^{i}_{j}}}}
	\label{eq:MA}
\end{equation}

By element-wise multiplying input feature map $ f_{k} $ with the corresponding weight map $ m_{k} $ and summing all the modalities, we can obtain a fused feature map $ f_{s} $ as:
\begin{equation}
	f_{s}= \sum_{k\in K}{f_{k}\cdot m_{k}}
	\label{eq:generation}
\end{equation}

Since the sum of $ m^{i}_{1}, \dots m^{i}_{|K|}$  is 1, the value range of fused feature representation $ f_{s} $ remains stable to improve the robustness for variable input modalities. Moreover, the relative sizes of $ v^{i}_{1}, \dots v^{i}_{|K|}$ (contain the latent multi-modal correlations learned from the CE module) are retained in the corresponding weights. In particular, when only one modality is available, all the values of the weight map are 1, which means $ f_{s} = f_{k} $ ($ k \in K $, $ |K|=1 $). In this case, the input feature representation remains unchanged. It enables the backbone network (the upstream processing model and the downstream decision model) to enhance its capability to encode and decode information from different modalities rather than relying on a particular one. It is crucial for variable multimodal data analysis.

\section{Experiments and Results}\label{experiment}
\subsection{Datasets}
\textbf{SHL2019.} The SHL (Sussex-Huawei Locomotion) Challenge 2019 \cite{shl:19} dataset provides data from seven sensors of a smartphone to recognize eight modes of locomotion and transportation (activities), including \textit{still}, \textit{walking}, \textit{run}, \textit{bike}, \textit{car}, \textit{bus}, \textit{train}, and \textit{subway}. The sensor data are collected from smartphones of a person with four locations, including the bag, trousers front pocket, breast pocket and hand. Each location is called ``Bag'', ``Hips'', ``Torso'', and ``Hand'', respectively. Data acquired from the locations except the ``Hand'' are given in the train subset, while the validation subset provides the data of all four locations. In the test subset, only unlabeled ``Hand'' location data are available.

\begin{figure}[tp]
	\centerline{\includegraphics[width=0.85\textwidth]{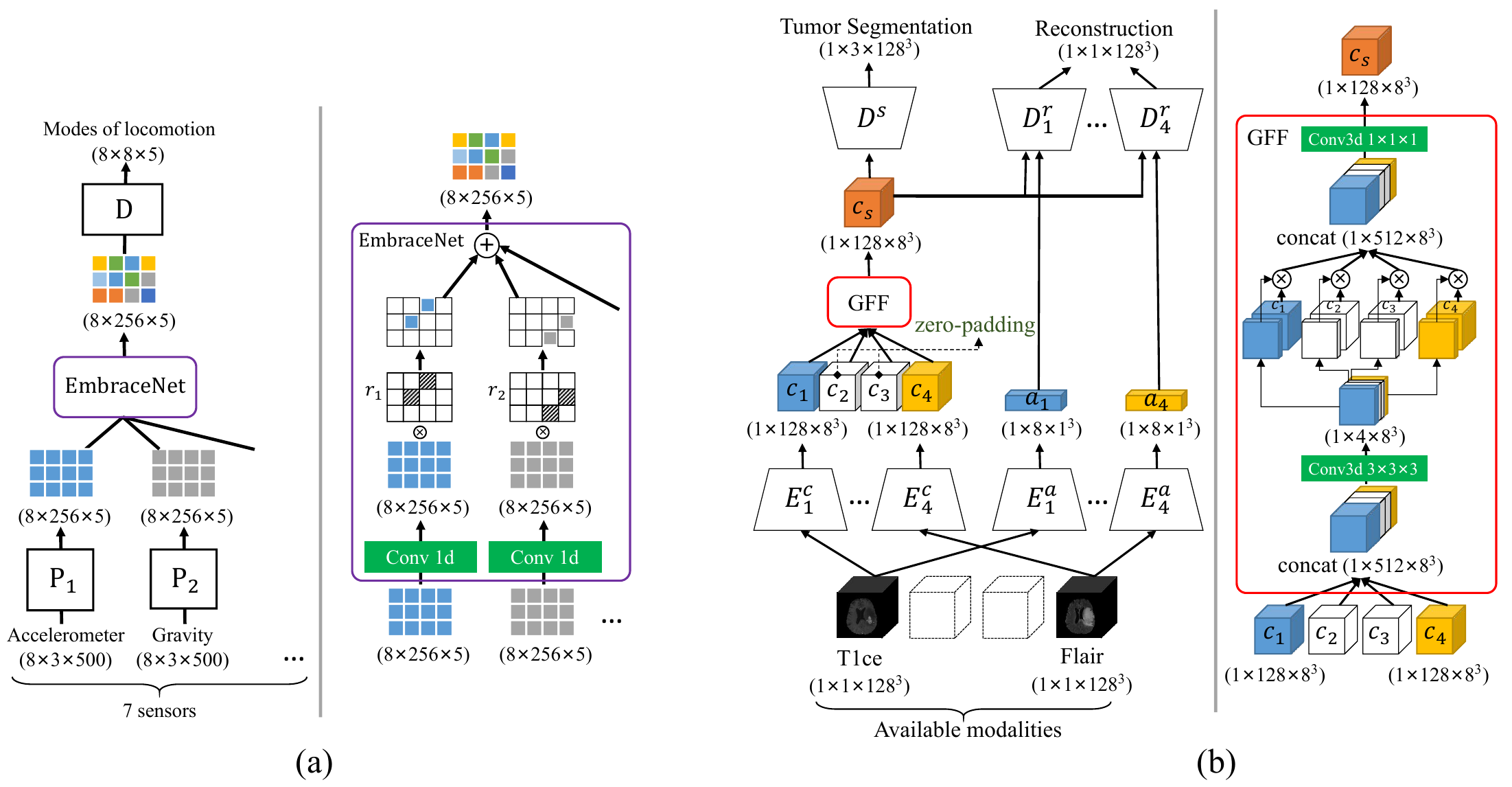}}
	\caption{(a) Activity recognition with EmbraceNet; (b) Bran tumor segmentation with GFF. (B $\times$ C $\times$ $ R_{f} $) is given, where B, C and $ R_{f} $ denotes the batch size, channels and data shape, respectively.}
	\label{fig4}
\end{figure}

\textbf{BraTS2020.} The BraTS2020 \cite{brats:20} dataset provide four modality scans: T1ce, T1, T2, FLAIR for brain tumor segmentation. It contains 369 subjects. To better represent the clinical application tasks, there are three mutually inclusive tumor regions: the enhancing tumor (ET), the tumor core (TC), and the whole tumor (WT)~\cite{brats:20}. We select 70\% data as training data, while 10\% and 20\% as validation and test data respectively. To prevent overfitting, two data augmentation techniques (randomly flip the axes and rotate with a random angle in $ [-10^{\circ}, 10^{\circ}] $) are applied during training. We apply z-score normalization \cite{ipk:18} to the volumes individually and randomly crop $ 128 \times 128 \times 128 $ patches as inputs to the networks.

\subsection{Baseline Methods} 
\textbf{EmbraceNet.} In the experiments on activity recognition, we compare SFusion with EmbraceNet \cite{cl2:19}, which employs a selection strategy (shown in Fig.~\ref{fig1} (b)) by generating feature masks ($ r_{1}, r_{2}, \dots, r_{7} $) with the rule of giving equal chances to all available modalities during each value selection. For a fair comparison, as shown in Fig.~\ref{fig4} (a), we adopt the same processing (P) and decision (D) model as used in \cite{cl2:19}. \textit{We obtain the performance of our fusion strategy by replacing EmbraceNet with SFusion.} Following \cite{cl2:19} setting, the batch size is set to 8. A cross-entropy loss and the Adam optimization method \cite{kb:14} with $ \beta_{1}= 0.9 $, $ \beta_{2}= 0.999 $ are employed. The learning rate is initially set to $ 1 \times 10^{-4} $ and reduced by a factor of 2 at every $1 \times 10^{5}$ steps. A total of $5 \times 10^{5}$ training steps are executed.

\textbf{GFF.} In the experiments on brain tumor segmentation, we compare SFusion with a gated feature fusion block (GFF) \cite{cdj:19}, which belongs to the convolution strategy (shown in Fig.~\ref{fig1}(c)). As shown in Fig.~\ref{fig4} (b), a feature disentanglement architecture is employed. Multimodal medical images are decomposed into the modality-invariant content and the modality-specific appearance code by encoders $ E^{c} $ and $ E^{a} $, respectively. The content codes (e.g., $ c_{2} $ and $ c_{3} $, shown in Fig.~\ref{fig4} (b)) of missing modalities are simulated with zero values. Then, all content codes are fused into a shared representation $ c_{s} $ by GFF. Given $ c_{s} $ , the tumor segmentation results are generated by the decoder $ D^{s} $. For a fair comparison, we adopt the same encoders ($ E^{c}_{i} $ and $ E^{a}_{i} $) and decoders ($ D^{s} $ and $ D^{r}_{i} $) as used in \cite{cdj:19}. \textit{We obtain the performance of our fusion strategy by replacing GFF with SFusion and removing the zero-padding operation.} The training max\_epoch is set to 200. Following \cite{cdj:19} setting, the batch size is set to 1. Adam \cite{kb:14} is utilized with a learning rate of $ 1\times 10^{-4} $ and progressively multiplies it by (1 - epoch / max\_epoch$)^{0.9}$. Losses of $ \mathcal{L}_{KL} $, $ \mathcal{L}_{rec} $ and $ \mathcal{L}_{seg} $ are employed as \cite{cdj:19}. During training, to simulate real missing modalities scenarios, each training patient's data is fixed to one of 15 possible missing cases. For a comprehensive evaluation, we test the performance of all 15 cases for each test patient.

Our implementations are on an NVIDIA RTX 3090(24G) with PyTorch 1.8.1.


\begin{table*}[b!]
	\centering
	\scriptsize 
	\renewcommand{\arraystretch}{1}
	\caption{Dice(\%) performance for MRI modalities being either absent ($\circ$) or present ($\bullet$). * denotes significant improvement provided by a Wilcoxon test ($ p$-values $ < 0.05 $).}
	\setlength{\tabcolsep}{2mm}{\begin{tabular}{cccc|cc|cc|cc}
			\toprule[1.4pt]
			\multicolumn{4}{c|}{Modalities} &
			\multicolumn{2}{c|}{WT} & \multicolumn{2}{c|}{TC} & \multicolumn{2}{c}{ET} \\
			\hline
			T1ce & T1 & T2 & Flair & GFF & SFusion & GFF & SFusion & GFF & SFusion  \\
			
			$\bullet$ &   $\circ$ &   $\circ$ &   $\circ$ & 68.24 & \textbf{69.75*} & 73.27 & \textbf{75.63*} & 69.30 & \textbf{71.94*} \\
			$\circ$   & $\bullet$ &   $\circ$ &   $\circ$ & 64.45 & \textbf{69.11*} & 46.93 & \textbf{53.86*} & 23.74 & \textbf{29.71*} \\
			$\circ$   &   $\circ$ & $\bullet$ &   $\circ$ &  \textbf{79.78} & 79.61 & 58.27 & \textbf{61.99*} & \textbf{36.13} & 35.87 \\
			$\circ$   &   $\circ$ &   $\circ$ & $\bullet$ &  81.82 & \textbf{83.97} & 50.53 & \textbf{52.84} & 29.50 & \textbf{34.40*} \\
			$\bullet$ & $\bullet$ &   $\circ$ &   $\circ$ &  74.99 & \textbf{75.30} & 75.89 & \textbf{80.35*} & 72.09 & \textbf{74.90*} \\
			$\bullet$ &   $\circ$ & $\bullet$ &   $\circ$ &  83.93 & \textbf{84.27*} & 79.55 & \textbf{81.48*} & 72.87 & \textbf{74.74*} \\
			$\bullet$ &   $\circ$ &   $\circ$ & $\bullet$ &  \textbf{87.34} & 87.32 & 79.01 & \textbf{79.06} & 74.89 & \textbf{75.82} \\
			$\circ$   & $\bullet$ & $\bullet$ &   $\circ$ &  81.76 & \textbf{81.78} & 59.75 & \textbf{66.67*} & 36.50 & \textbf{40.38*} \\
			$\circ$   & $\bullet$ &   $\circ$ & $\bullet$ &  85.86 & \textbf{86.39} & 61.92 & \textbf{62.31} & 37.52 & \textbf{38.22} \\
			$\circ$   &   $\circ$ & $\bullet$ & $\bullet$ &  86.99 & \textbf{87.50*} & 61.92 & \textbf{66.38*} & 38.94 & \textbf{41.46*} \\
			$\bullet$ & $\bullet$ & $\bullet$ &   $\circ$ & 84.48 & \textbf{84.59} & 79.83 & \textbf{82.32*} & 73.74 & \textbf{74.78} \\
			$\bullet$ & $\bullet$ &   $\circ$ & $\bullet$ & 88.03 & \textbf{88.04} & 80.50 & \textbf{82.04*} & 74.53 & \textbf{75.44} \\
			$\bullet$ &   $\circ$ & $\bullet$ & $\bullet$ & 88.75 & \textbf{89.11*} & 81.60 & \textbf{82.06} & 74.43 & \textbf{74.91} \\
			$\circ$   & $\bullet$ & $\bullet$ & $\bullet$ & 86.84 & \textbf{87.63*} & 65.38 & \textbf{68.76*} & 40.90 & \textbf{43.53*} \\
			$\bullet$ & $\bullet$ & $\bullet$ & $\bullet$ & 88.65 & \textbf{88.93} & 81.29 & \textbf{82.18} & \textbf{74.55} & 73.76 \\
			\hline
			\multicolumn{4}{c|}{Average}     & 82.13 & \textbf{82.89*} & 69.04 & \textbf{71.86*} & 55.31 & \textbf{57.32*} \\
			\bottomrule[1.2pt]
	\end{tabular}}%
	\label{tab1}%
\end{table*}%

\subsection{Results.}
\textbf{Activity recognition.} We compare SFusion with the EmbraceNet \cite{cl2:19} on SHL2019. As shown in Table~\ref{tab3}, we also compare the results of other fusion methods, which use the same processing (P) model and decision (D) model as \cite{cl2:19}. (1) In the early fusion method, the data of seven sensors are concatenated along their $ C $ dimension. The prediction results are obtained by inputting the concatenation into a network of P and D in series. (2) For the intermediate fusion approach, the EmbraceNet is replaced with the concatenation of feature representations along their $ R_{f} $ dimension. (3) In the late fusion method, an independent network of P and D in series is trained for each sensor, and then the decision is made from the averaged softmax outputs. (4) In the confidence fusion model, the EmbraceNet is replaced with the confidence calculation and fusion layers in \cite{cl:18}. The results of different fusion methods on the validation data are presented in Table~\ref{tab3}. Our proposed SFusion outperforms the EmbraceNet in all four smartphone locations and improves the overall accuracy from 65.22\% to 67.47\%.

\begin{figure}[t]
	\begin{minipage}[t]{0.4\textwidth}
		\centering
		\scriptsize
		\renewcommand{\arraystretch}{1.1}
		\makeatletter\def\@captype{table}\makeatother\caption{Ablation experiments.}
		\begin{tabular}{c|c|c|c}
			\hline
			Dice(\%)& $ w/o $ CE & $ w/o $ MA & SFusion
			\\
			\hline
			WT& 82.42 & 82.76 & \textbf{82.89}\\
			TC& 70.39 & 70.93 & \textbf{71.86}\\
			ET & 55.65 & 55.56 & \textbf{57.32} \\
			\hline
		\end{tabular}
		\label{tab2}%
	\end{minipage}
	\begin{minipage}[t]{0.6\textwidth}
		\centering
		\scriptsize
		\renewcommand{\arraystretch}{1.1}
		\makeatletter\def\@captype{table}\makeatother\caption{ $ \dag $ denotes results from \cite{ddn:22}.}
		\setlength{\tabcolsep}{1.5mm}{\begin{tabular}{l|ccccc}
				\hline
				Dice(\%) & WT & TC & ET & Overall \\
				\hline
				U-HVED$ \dag $ \cite{djm:19} & 75.8 & 63.2 & 40.7 & 59.9 \\
				ACNet$ \dag $ \cite{wzl:21}  & 52.5 & 46.9 & 41.8 & 47.1 \\
				D$^{2}$-Net$ \dag $ \cite{ddn:22} & 76.2 & 66.5 & 42.3 & 61.7 \\
				SF\_FDGF                     & \textbf{82.1} & \textbf{69.2} & \textbf{54.9} &	\textbf{68.7}  \\
				\hline
				
		\end{tabular}}
		\label{tab4}%
	\end{minipage}
\end{figure}

\textbf{Brain tumor segmentation.} The quantitative segmentation results are shown in Table~\ref{tab1}. Compared with GFF, the network integrated with SFusion achieves better average performance over the 15 possible combinations in all three tasks. In particular, SFusion outperforms GFF for all the possible combinations in TC segmentation. Overall, SFusion achieves better Dice scores in most situations (13,15,13 situations for WT, TC and ET segmentation, respectively). In addition, we conduct the statistical significance analysis. The number of situations with significant improvement are 6, 10 and 8 for WT, TC and ET, respectively. It is provided by a Wilcoxon test ($ p$-values $ < 0.05 $). Besides, we find no significant drop in performance caused by SFusion. In addition, we compare the SF\_FDGF (where GFF is replaced by SFusion) with current state-of-the-art methods. Table~\ref{tab4} presents the average dice of 15 situations. For a fair comparison, we conduct experiments on BraTS2018, adopt the same data partition as~\cite{ddn:22}, and cite the results in~\cite{ddn:22}. SF\_FDGF achieves the best performance and verifies the effectiveness of the SFusion. 

\textbf{Ablation experiments}.The correlation extraction (CE) module and the modal attention (MA) module are two key components in SFusion. We evaluate the SFusion without CE and MA, respectively. SFusion without CE denotes that feature representations are directly fed into the MA module (Fig.~\ref{fig2}). SFusion without MA means that we directly add the calculated feature representations ($ I^{\prime} $) up to get the fusion result. As shown in Table~\ref{tab3}, we can find that SFusion without CE performs worse than other methods. Compared with EmbraceNet, the improvement of SFusion without MA is inconspicuous. As shown in Table.~\ref{tab2}, we present the averaged performance over the 15 possible combinations on BraTS2020. It shows that both the CE and MA module lead to performance improvement across all the tumor regions. Therefore, ablation experiments on two different tasks show that both CE and MA play an important role in SFusion.

%
%

\section{Conclusion}
In this paper, we propose a self-attention based N-to-One fusion block SFusion to tackle the problem of multimodal missing modalities fusion. As a data-dependent fusion strategy, SFusion can automatically learn the latent correlations between different modalities and builds a shared feature representation. The entire fusion process is based on available data without simulating missing modalities. In addition, SFusion has compatibility with any kind of upstream processing model and downstream decision model, making it universally applicable to different tasks. We show that it can be integrated into existing backbone networks by replacing their fusion operation or block to improve activity recognition and achieve brain tumor segmentation performance. In particular, by integrating with SFusion, SF\_FDGF achieves the state-of-the-art performance. In the future, we will explore other tasks related to variable multimodal fusion with SFusion.

\subsubsection{Acknowledgements} This work is supported in part by the Guangdong Provincial Natural Science Foundation (2023A1515011431), the Guangzhou Science and Technology Planning Project (202201010092), the National Natural Science Foundation of China (72074105), NSF-1850492 and NSF-2045804.

%
%
%
\bibliographystyle{splncs04}
\bibliography{sfusion}

\begin{thebibliography}{10}
\providecommand{\url}[1]{\texttt{#1}}
\providecommand{\urlprefix}{URL }
\providecommand{\doi}[1]{https://doi.org/#1}

\bibitem{brats:20}
Bakas, S., Menze, B., Davatzikos, C., Kalpathy-Cramer, J., Farahani, K.,
  et~al.: {MICCAI Brain Tumor Segmentation (BraTS) 2020 Benchmark: "Prediction
  of Survival and Pseudoprogression"} (Mar 2020). \doi{10.5281/zenodo.3718904}

\bibitem{cjg:18}
Chartsias, A., Joyce, T., Giuffrida, M.V., Tsaftaris, S.A.: Multimodal mr
  synthesis via modality-invariant latent representation. IEEE Transactions on
  Medical Imaging  \textbf{37}(3),  803--814 (2018).
  \doi{10.1109/TMI.2017.2764326}

\bibitem{csc:13}
Chavarriaga, R., Sagha, H., Calatroni, A., Digumarti, S.T., Tr{\"o}ster, G.,
  Mill{\'a}n, J.d.R., Roggen, D.: The opportunity challenge: A benchmark
  database for on-body sensor-based activity recognition. Pattern Recognition
  Letters  \textbf{34}(15),  2033--2042 (2013)

\bibitem{cjk:15}
Chen, C., Jafari, R., Kehtarnavaz, N.: Utd-mhad: A multimodal dataset for human
  action recognition utilizing a depth camera and a wearable inertial sensor.
  In: 2015 IEEE International conference on image processing (ICIP). pp.
  168--172. IEEE (2015)

\bibitem{cdj:19}
Chen, C., Dou, Q., Jin, Y., Chen, H., Qin, J., Heng, P.A.: Robust multimodal
  brain tumor segmentation via feature disentanglement and gated fusion. In:
  International Conference on Medical Image Computing and Computer-Assisted
  Intervention. pp. 447--456. Springer (2019)

\bibitem{cdjlh:21}
Chen, C., Dou, Q., Jin, Y., Liu, Q., Heng, P.A.: Learning with privileged
  multimodal knowledge for unimodal segmentation. IEEE Transactions on Medical
  Imaging pp.~1--1 (2021). \doi{10.1109/TMI.2021.3119385}

\bibitem{cl:18}
Choi, J.H., Lee, J.S.: Confidence-based deep multimodal fusion for activity
  recognition. In: Proceedings of the 2018 ACM International Joint Conference
  and 2018 International Symposium on Pervasive and Ubiquitous Computing and
  Wearable Computers. pp. 1548--1556 (2018)

\bibitem{cl:19}
Choi, J.H., Lee, J.S.: Embracenet: A robust deep learning architecture for
  multimodal classification. Information Fusion  \textbf{51},  259--270 (2019)

\bibitem{cl2:19}
Choi, J.H., Lee, J.S.: Embracenet for activity: A deep multimodal fusion
  architecture for activity recognition. In: Adjunct Proceedings of the 2019
  ACM International Joint Conference on Pervasive and Ubiquitous Computing and
  Proceedings of the 2019 ACM International Symposium on Wearable Computers.
  pp. 693--698 (2019)

\bibitem{djm:19}
Dorent, R., Joutard, S., Modat, M., Ourselin, S., Vercauteren, T.: Hetero-modal
  variational encoder-decoder for joint modality completion and segmentation.
  In: Shen, D., Liu, T., Peters, T.M., Staib, L.H., Essert, C., Zhou, S., Yap,
  P.T., Khan, A. (eds.) Medical Image Computing and Computer Assisted
  Intervention -- MICCAI 2019. pp. 74--82. Springer International Publishing,
  Cham (2019)

\bibitem{gm:13}
Graves, M.J., Mitchell, D.G.: Body mri artifacts in clinical practice: a
  physicist's and radiologist's perspective. Journal of Magnetic Resonance
  Imaging  \textbf{38}(2),  269--287 (2013)

\bibitem{glh:19}
Guo, Z., Li, X., Huang, H., Guo, N., Li, Q.: Deep learning-based image
  segmentation on multimodal medical imaging. IEEE Transactions on Radiation
  and Plasma Medical Sciences  \textbf{3}(2),  162--169 (2019)

\bibitem{hgc:16}
Havaei, M., Guizard, N., Chapados, N., Bengio, Y.: Hemis: Hetero-modal image
  segmentation. In: Ourselin, S., Joskowicz, L., Sabuncu, M.R., Unal, G.,
  Wells, W. (eds.) Medical Image Computing and Computer-Assisted Intervention
  -- MICCAI 2016. pp. 469--477. Springer International Publishing, Cham (2016)

\bibitem{hmz:20}
Hu, M., Maillard, M., Zhang, Y., Ciceri, T., La~Barbera, G., Bloch, I., Gori,
  P.: Knowledge distillation from multi-modal to mono-modal segmentation
  networks. In: Martel, A.L., Abolmaesumi, P., Stoyanov, D., Mateus, D.,
  Zuluaga, M.A., Zhou, S.K., Racoceanu, D., Joskowicz, L. (eds.) Medical Image
  Computing and Computer Assisted Intervention -- MICCAI 2020. pp. 772--781.
  Springer International Publishing, Cham (2020)

\bibitem{ipk:18}
Isensee, F., Petersen, J., Klein, A., Zimmerer, D., Jaeger, P.F., Kohl, S.,
  Wasserthal, J., Koehler, G., Norajitra, T., Wirkert, S., et~al.: nnu-net:
  Self-adapting framework for u-net-based medical image segmentation. arXiv
  preprint arXiv:1809.10486  (2018)

\bibitem{kb:14}
Kingma, D.P., Ba, J.: Adam: A method for stochastic optimization. arXiv
  preprint arXiv:1412.6980  (2014)

\bibitem{las:19}
Lau, K., Adler, J., Sj{\"o}lund, J.: A unified representation network for
  segmentation with missing modalities. arXiv preprint arXiv:1908.06683  (2019)

\bibitem{nkk:11}
Ngiam, J., Khosla, A., Kim, M., Nam, J., Lee, H., Ng, A.Y.: Multimodal deep
  learning. In: ICML (2011)

\bibitem{oap:21}
Ouyang, J., Adeli, E., Pohl, K.M., Zhao, Q., Zaharchuk, G.: Representation
  disentanglement for multi-modal brain mri analysis. In: International
  Conference on Information Processing in Medical Imaging. pp. 321--333.
  Springer (2021)

\bibitem{szw:21}
Shen, L., Zhu, W., Wang, X., Xing, L., Pauly, J.M., Turkbey, B., Harmon, S.A.,
  Sanford, T.H., Mehralivand, S., Choyke, P.L., Wood, B.J., Xu, D.:
  Multi-domain image completion for random missing input data. IEEE
  Transactions on Medical Imaging  \textbf{40}(4),  1113--1122 (2021).
  \doi{10.1109/TMI.2020.3046444}

\bibitem{vsp:17}
Vaswani, A., Shazeer, N., Parmar, N., Uszkoreit, J., Jones, L., Gomez, A.N.,
  Kaiser, L.u., Polosukhin, I.: Attention is all you need. In: Guyon, I.,
  Luxburg, U.V., Bengio, S., Wallach, H., Fergus, R., Vishwanathan, S.,
  Garnett, R. (eds.) Advances in Neural Information Processing Systems.
  vol.~30. Curran Associates, Inc. (2017)

\bibitem{shl:19}
Wang, L., Gjoreski, H., Ciliberto, M., Mekki, S., Valentin, S., Roggen, D.:
  Enabling reproducible research in sensor-based transportation mode
  recognition with the sussex-huawei dataset. IEEE Access  \textbf{7},
  10870--10891 (2019)

\bibitem{wzl:21}
Wang, Y., Zhang, Y., Liu, Y., Lin, Z., Tian, J., Zhong, C., Shi, Z., Fan, J.,
  He, Z.: Acn: Adversarial co-training network for brain tumor segmentation
  with missing modalities. In: de~Bruijne, M., Cattin, P.C., Cotin, S., Padoy,
  N., Speidel, S., Zheng, Y., Essert, C. (eds.) Medical Image Computing and
  Computer Assisted Intervention -- MICCAI 2021. pp. 410--420. Springer
  International Publishing, Cham (2021)

\bibitem{ddn:22}
Yang, Q., Guo, X., Chen, Z., Woo, P.Y., Yuan, Y.: D2-net: Dual disentanglement
  network for brain tumor segmentation with missing modalities. IEEE
  Transactions on Medical Imaging  (2022)

\bibitem{lcr:21}
Zhou, T., Canu, S., Vera, P., Ruan, S.: Latent correlation representation
  learning for brain tumor segmentation with missing mri modalities. IEEE
  Transactions on Image Processing  \textbf{30},  4263--4274 (2021)

\end{thebibliography}

\end{document}